\documentclass[table]{IEEEtran}
\usepackage[utf8]{inputenc}
\usepackage{array}
\usepackage{wrapfig}
\usepackage{multirow}
\usepackage{tabu}
\usepackage{xcolor}
\usepackage{cite}
\usepackage{amsmath,amssymb,amsfonts}
\usepackage{algorithmic}
\usepackage{algorithm}
\usepackage{graphicx}
\usepackage{textcomp}
\usepackage{caption}
\usepackage{comment}
\usepackage{subcaption}
\usepackage{float}
\usepackage{eucal}
\usepackage{soul}
\usepackage{fixltx2e}
\usepackage{enumerate}
\usepackage{cuted}
\usepackage{tabularx}
\usepackage{pgf-pie}
\usepackage{enumitem}
\usepackage{flushend}
\def\BibTeX{{\rm B\kern-.05em{\sc i\kern-.025em b}\kern-.08em
    T\kern-.1667em\lower.7ex\hbox{E}\kern-.125emX}}
\usepackage{fancyhdr}

\fancypagestyle{plain}{
  \fancyhf{}
  \fancyhead[C]{}     
  \fancyfoot[L]{This paper has been accepted at the IEEE International Symposium on Circuits and Systems (ISCAS) to be held in May 2025 at London, UK.}

}
\usepackage{eso-pic}
\newcommand*\darkgreycircled[1]{\tikz[baseline=(char.base)]{
            \node[shape=circle,fill=darkgray,text=white,draw=black,inner sep=0.1pt] (char) {#1};}}
            
\begin{document}
\title{Neural Precision Polarization: Simplifying Neural Network Inference with Dual-Level Precision}
\author{Dinithi Jayasuriya, Nastaran Darabi, Maeesha Binte Hashem, and Amit Ranjan Trivedi \\
University of Illinois at Chicago, IL, USA, email: \{dkasth2, ndarab2, mbinte2, amitrt\}@uic.edu
\vspace{-1.75em}}

\maketitle
\begin{abstract}
We introduce a \textit{precision polarization} scheme for DNN inference that utilizes only very low and very high precision levels, assigning low precision to the majority of network weights and activations while reserving high precision paths for targeted error compensation. This separation allows for distinct optimization of each precision level, thereby reducing memory and computation demands without compromising model accuracy. In the discussed approach, a floating-point model can be trained in the cloud and then downloaded to an edge device, where network weights and activations are directly quantized to meet the edge devices' desired level (such as NF4/INT8). To address accuracy loss from quantization, surrogate paths are introduced, leveraging low-rank approximations on a layer-by-layer basis. These paths are trained with a sensitivity-based metric on minimal training data to recover accuracy loss under quantization as well as due to process variability such as when the main prediction path is implemented using analog acceleration. Our simulation results show that neural precision polarization enables $\sim$464 TOPS/W MAC efficiency and reliability by integrating rank-8 error recovery paths with highly efficient, though potentially unreliable, bitplane-wise compute-in-memory processing.
\end{abstract}

\begin{IEEEkeywords}
Edge computing, Low-Ranked Approximation, Compute-in-Memory
\end{IEEEkeywords}

\section{Introduction}
\label{sec:introduction}
\textit{Intelligent edge computing} is increasingly in demand due to the rise of edge devices for allowing prediction models to process data locally at the user-environment interface. This shift enables real-time decisions, reduces reliance on cloud resources, and improves response times by minimizing network dependency. However, deploying large-scale DNNs at the edge faces constraints in computational power, memory, and energy\cite{qu2022enabling}. Low precision DNN processing mitigates these challenges by reducing input and weight bit-widths, which decreases data movement and arithmetic costs, and increases parallelism to improve computational efficiency.

Notably, the focus on low precision DNN inference also aligns well with emerging technologies like analog processing. For example, non-volatile memories (NVM) such as memristors, which store data as resistance states and perform analog computation, enable in-memory processing, eliminating the costly data movement inherent in von Neumann architectures; however, memristors can only reliably store limited bit precision per device. Likewise, analog processing in deep learning offers advantages by leveraging physical laws (e.g., Kirchhoff's law) to perform operations like addition directly in-circuit, bypassing digital arithmetic. These systems however remain efficient only at low precision, as higher precision increases signal control complexity and resource demand.

Despite gains in computational efficiency, reducing the precision of DNNs can significantly impact prediction accuracy due to the quantization of model parameters and activations. Lower bit-widths restrict the model’s capacity to capture fine-grained patterns, leading to performance degradation. Furthermore, at reduced precision, process variability can amplify quantization errors, compounding the problem. Prior works address precision scaling by using mixed precision inference, where layer-wise precision adapts to mitigate accuracy loss. However, supporting multiple precision levels on hardware significantly increases complexity, which eclipses benefits. 

\begin{figure}
    \centering
    \includegraphics[width=\columnwidth]{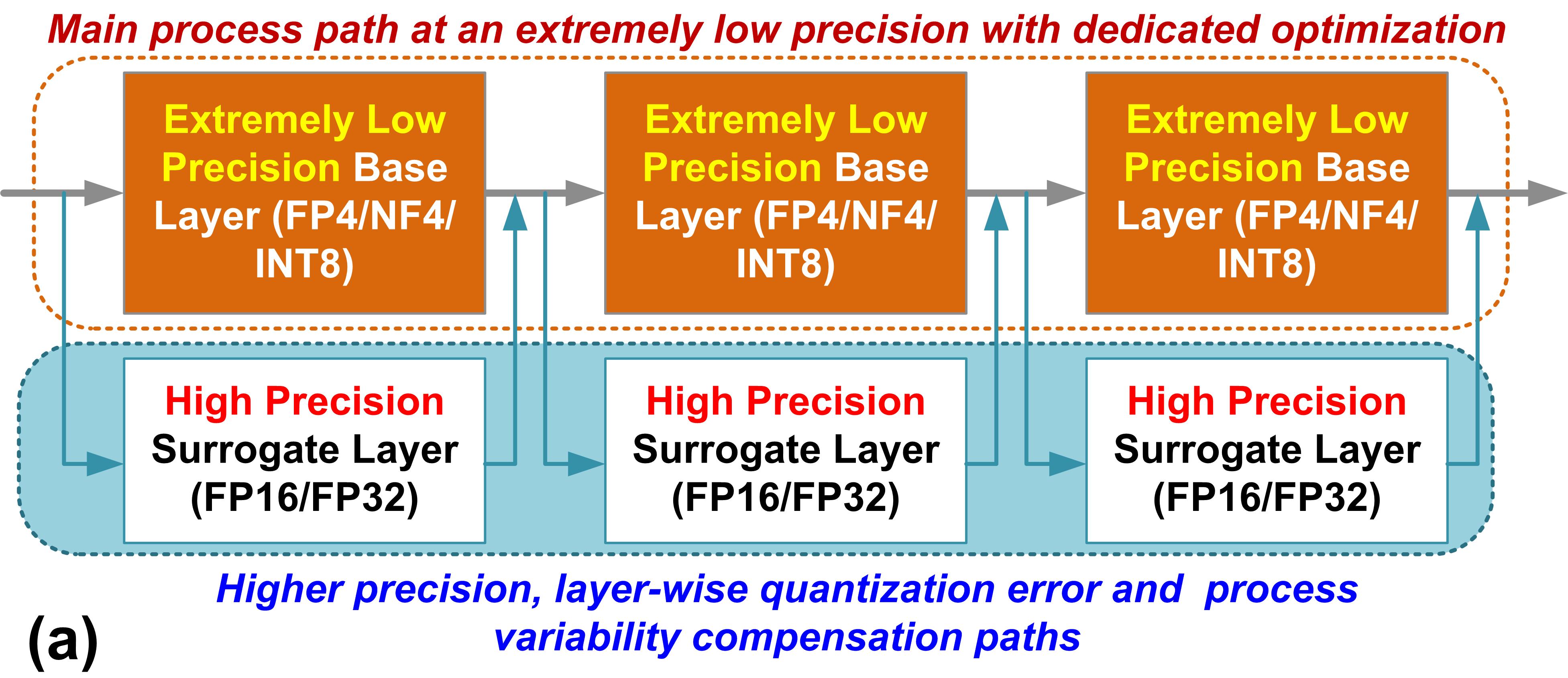}
    \includegraphics[width=\columnwidth]{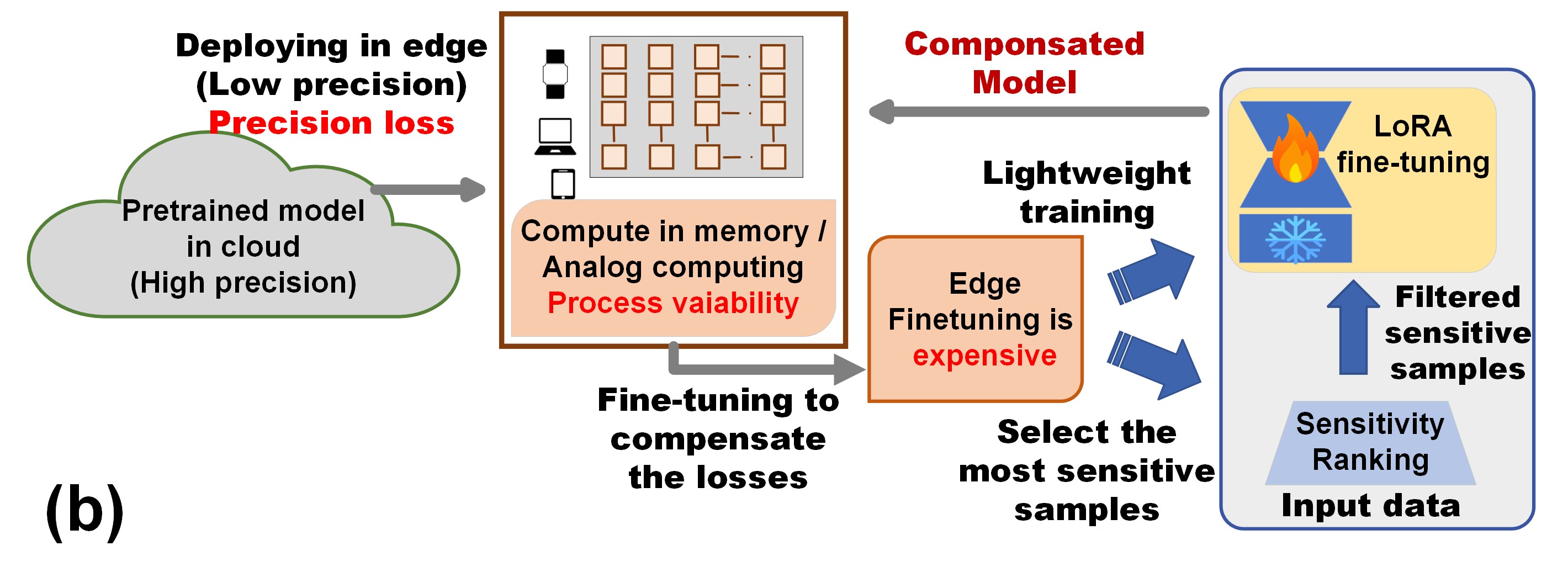}
\caption{\textbf{Overview of Neural Precision Polarization (NPP)}:\textbf{(a)} Neural precision polarization employs only two quantization levels, with most weights at ultra-low precision (e.g., FP4 or N4) and selective high-precision surrogate paths to mitigate accuracy loss. This dual-level approach enables dedicated optimization and simplified implementation of network inference. \textbf{(b)} Under NPP, a cloud-trained floating-point model is downloaded to the edge, with weights and activations quantized to meet the edge. Layer-wise surrogate paths with low-rank approximations and sensitivity-based metrics to recover accuracy with minimal overhead and retraining data. Compute-in-memory processing of low-rank paths is discussed. \vspace{-15pt}}
    \label{fig:framework}
\end{figure}

To address these challenges, we propose \textit{neural precision polarization (NPP)}, implementing processing with only two precision levels. In Fig. 1, under NPP the majority of network weights are processed in ultra-low precision (e.g., NF4 or FP4) to meet device constraints, with limited high-precision paths added to recover accuracy loss and handle precision variability and aging impacts. Towards such precision polarization, we present a novel approach that tackles them using a co-designed learning-inference scheme at all three desiderata:

\begin{itemize}[leftmargin=5pt, label=\textbullet, itemsep=0em]
\item \textit{Parametric Efficiency}: We employ a low-rank approximation of neural network, with most layer weights implemented at a very low precision and using energy-efficient but variability-prone technologies such as analog processing. To compensate the inference errors, we augment these layers with high-precision surrogate paths that contain much fewer parameters.
\item \textit{Sample Efficient Retraining}: Since our approach requires retraining to overcome device-dependent variability, to mitigate the retraining cost, we propose a sensitivity-metric to prioritize training data. Retraining with only sensitive data recovers the accuracy loss while minimizing the cost.
\item \textit{Energy Efficiency for Inference}: We discuss the integration of highly efficient, though potentially unreliable, bitplane-wise in-memory processing with the proposed low rank error recovery paths to achieve state-of-the-art energy efficiency and reliability on advanced networks.
\end{itemize} 

\section{Neural Precision Polarization}
Numerous studies have explored precision scaling and mixed-precision inference to reduce DNN energy by lowering bit-widths, minimizing power for data movement, memory access, and computations. \cite{lacey2018stochastic} showed that layer-wise precision scaling maximizes energy savings, applying lower precision to less critical layers while maintaining accuracy in key layers. \cite{reguero2025energy} introduced a mixed-precision edge inference framework with tensor-sliced learned precision and a hardware-aware cost function. Similarly, \cite{risso2022channel} proposed a NAS method with channel-level mixed precision, optimizing bit-widths for memory and computational efficiency. \cite{nasrin2022enos} demonstrated dynamic precision-scalable multipliers, nesting 2-bit or 4-bit multiplications within an 8-bit structure. Notably, adaptable and mixed-precision frameworks generally aim to integrate various precision levels within a single hardware architecture. In contrast, our \textit{neural precision polarization (NPP)} dedicates the main processing path to ultra-low precision (e.g., 4-bit or 8-bit), while high precision is reserved solely for surrogate quantization error and process variability compensation paths. By clearly distinguishing precision requirements, this approach enables targeted design optimizations.\vspace{-10pt}

\begin{figure}[t!]
    \centering
    \includegraphics[width=0.48\linewidth,height=0.479\linewidth]{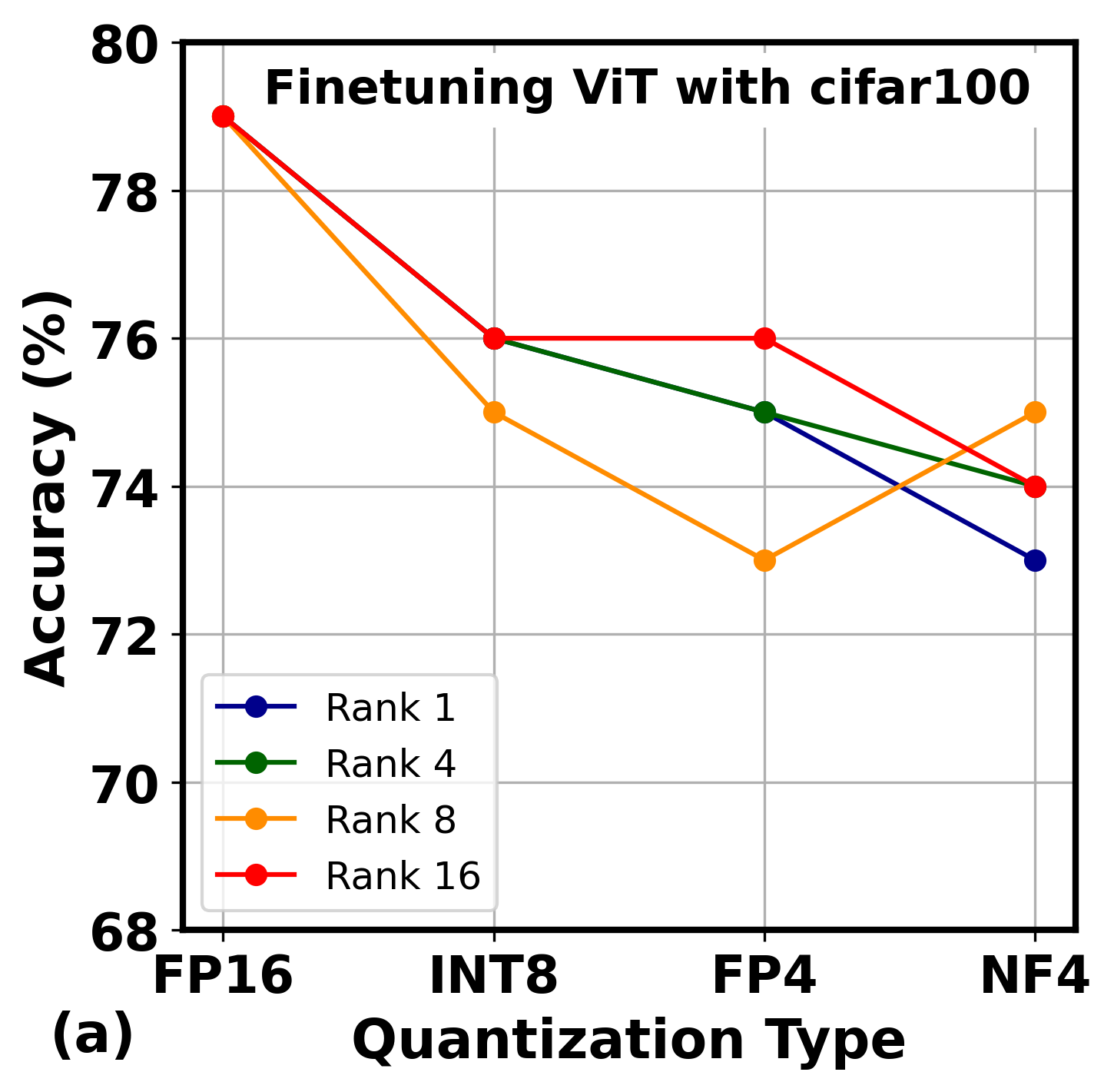}
    \includegraphics[width=0.48\linewidth,height=0.479\linewidth]{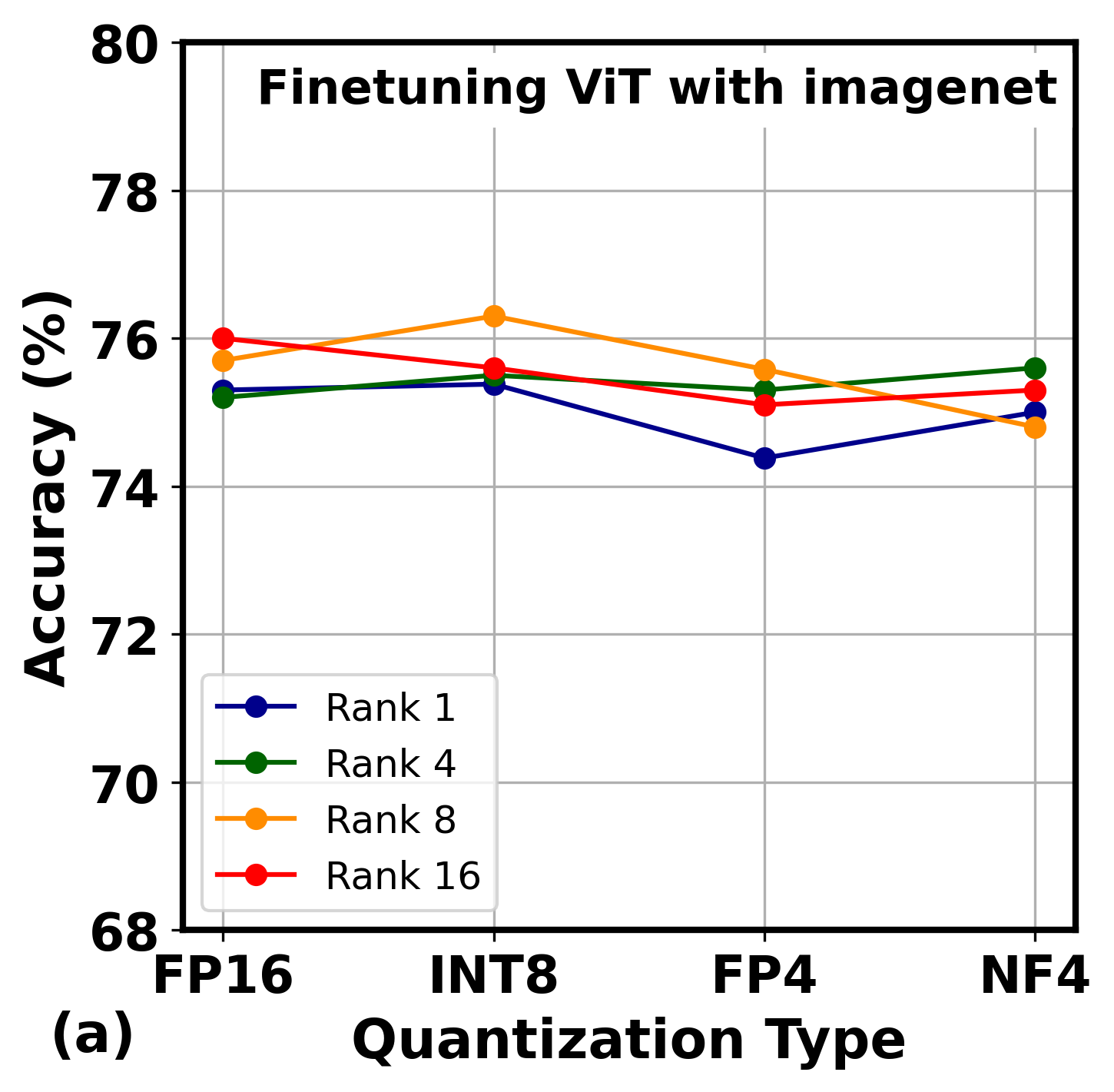}
    \caption{\textbf{Quantization Error Compensation with NPP}: LoRA fine-tuning recovers quantization accuracy loss in ViT models on (a) CIFAR100 and (b) ImageNet datasets, respectively with recovery varying by rank. \vspace{-15pt}}
    \label{fig:Quant}
\end{figure}

\subsection{Background: Low Rank Adaptation of DNNs}
NPP leverages a low-rank adaptation of neural network for precision polarization. Low-Rank Adaptation (LoRA) \cite{hu2021lora} is a fine-tuning method that updates a small subset of parameters in pretrained models, preserving accuracy while enhancing generalization to new tasks. Instead of updating the full weight matrix \( W \in \mathbb{R}^{D \times H} \) during fine-tuning, LoRA introduces two low-rank matrices \( A \in \mathbb{R}^{D \times r} \) and \( B \in \mathbb{R}^{r \times H} \) with \( r \ll \min(D, H) \), approximating the update as \( \Delta W = AB \). The new weight is then \( W_{\text{new}} = W + AB \), freezing \( W \) and only training \( A \) and \( B \). Among prior works, QLoRA~\cite{dettmers2024qlora} combines quantization with LoRA to fine-tune low-precision models, applying LoRA selectively to mitigate quantization-related accuracy loss. LoRA-the-Explorer (LTE)~\cite{huh2024training} introduces training from scratch with LoRA, optimizing for distributed settings. LoRA+~\cite{hayou2024lora+} addresses original LoRA limitations by introducing differential learning rates for low-rank matrices A and B, enhancing fine-tuning efficiency and performance while maintaining computational costs. While LoRA methods are mainly used in large language and multimodal models to reduce the cost of full model retraining on High-Performance Computing (HPC) systems, we exploit their effectiveness for precision polarization and process variability compensation.\vspace{-10pt}

\begin{figure}
    \centering    
    \includegraphics[width=0.48\linewidth,height=0.479\linewidth]{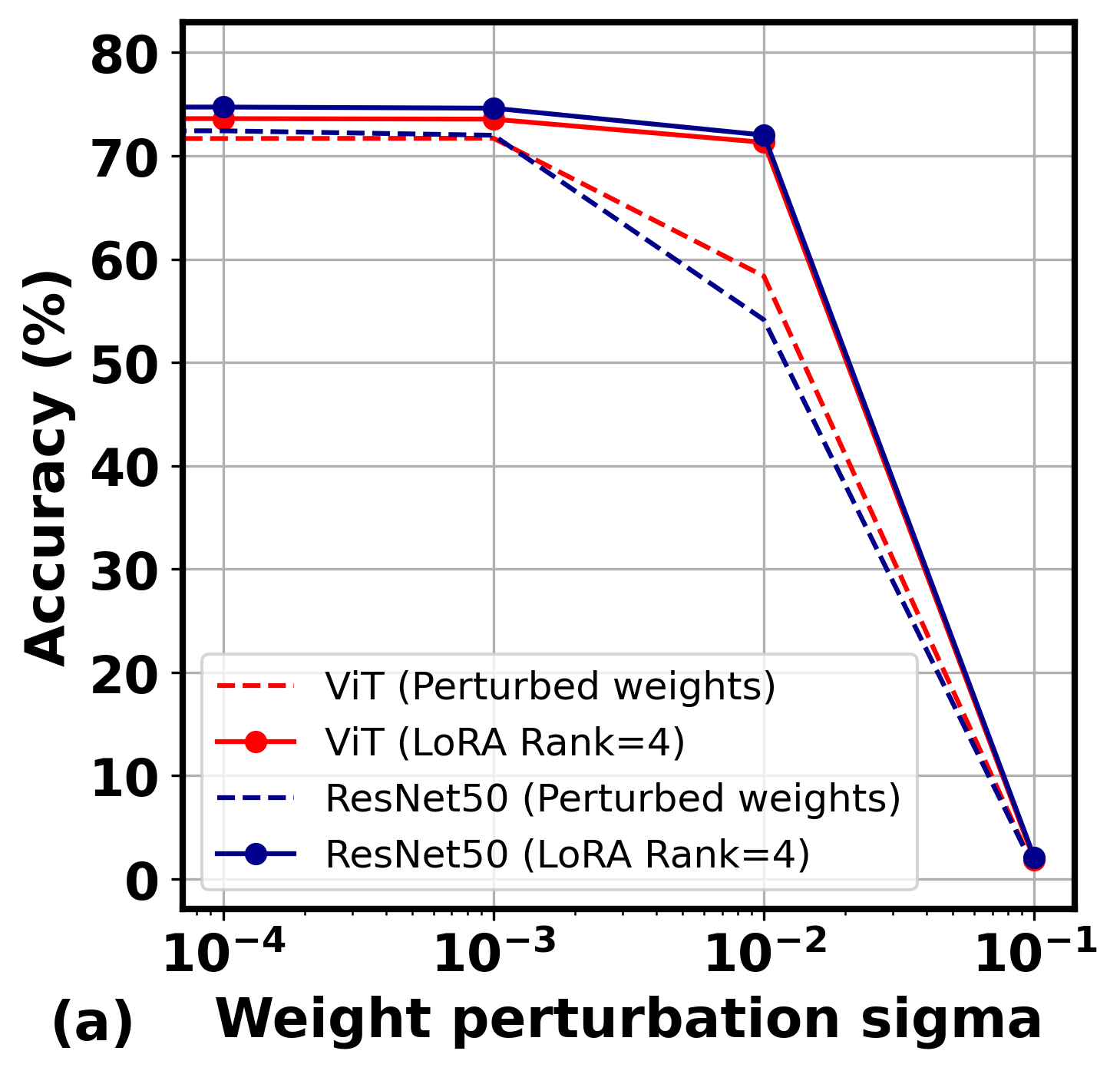}
    \includegraphics[width=0.48\linewidth,height=0.479\linewidth]{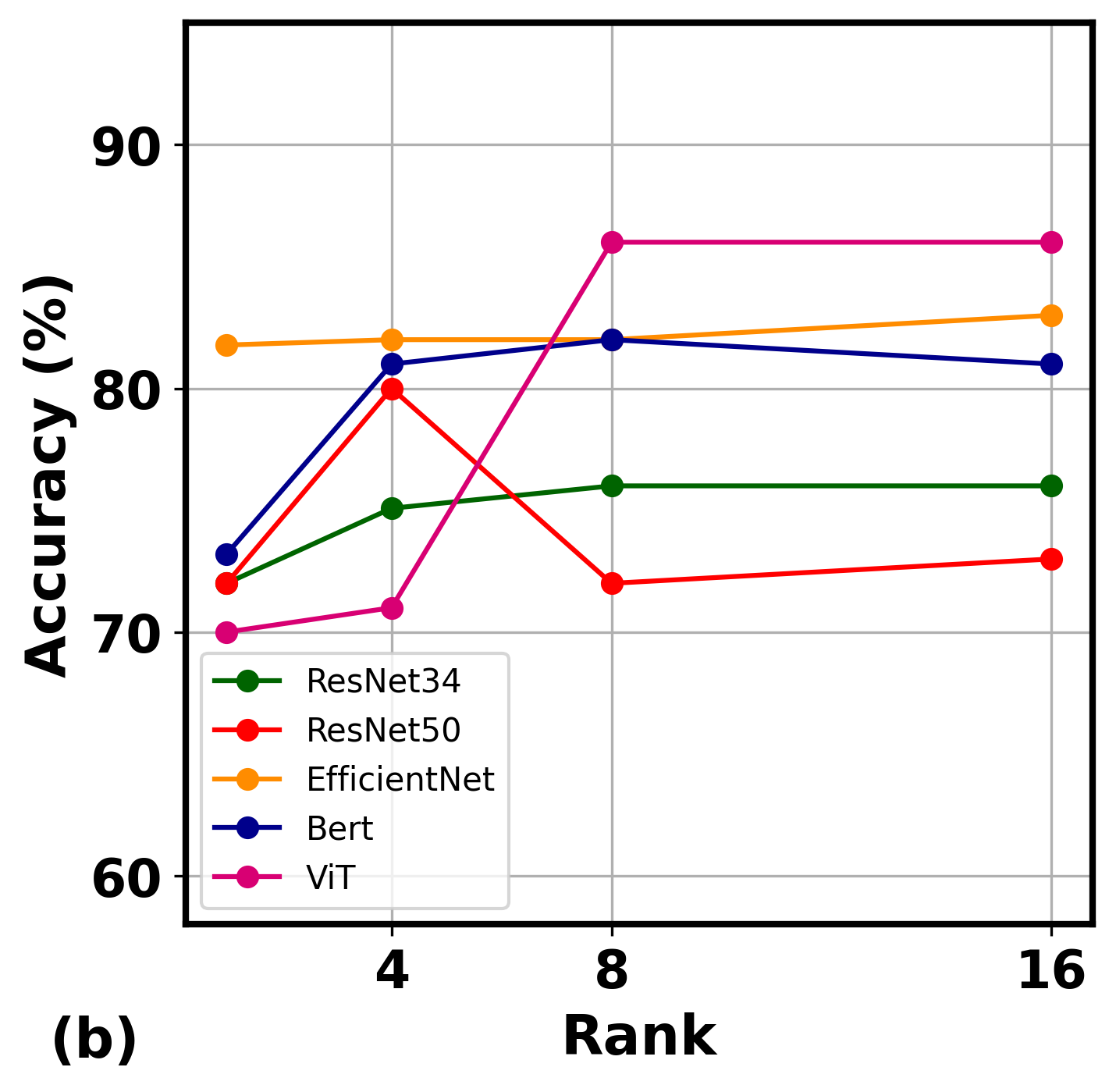}
    \includegraphics[width=0.48\linewidth,height=0.479\linewidth]{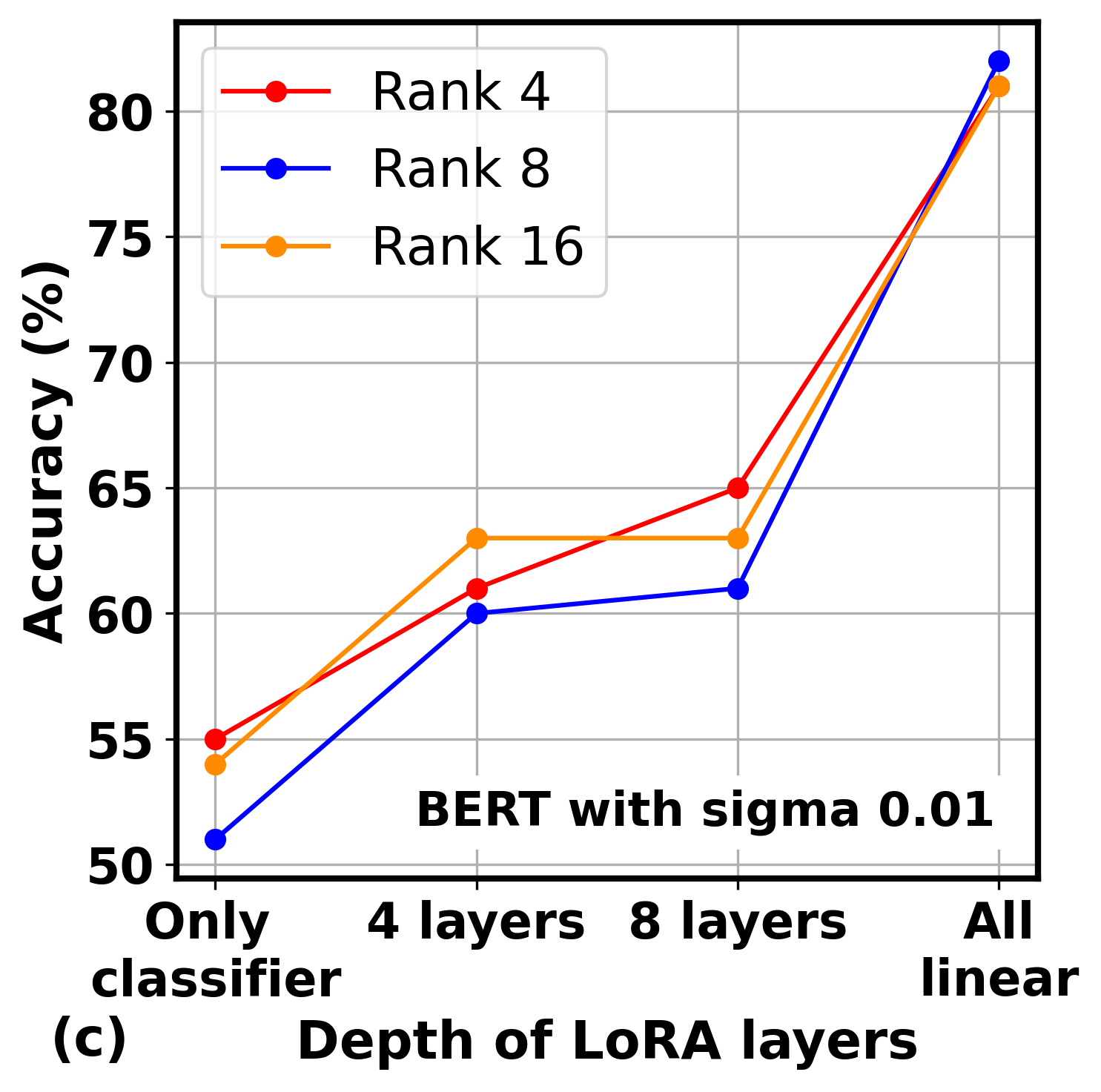}
    \includegraphics[width=0.48\linewidth,height=0.479\linewidth]{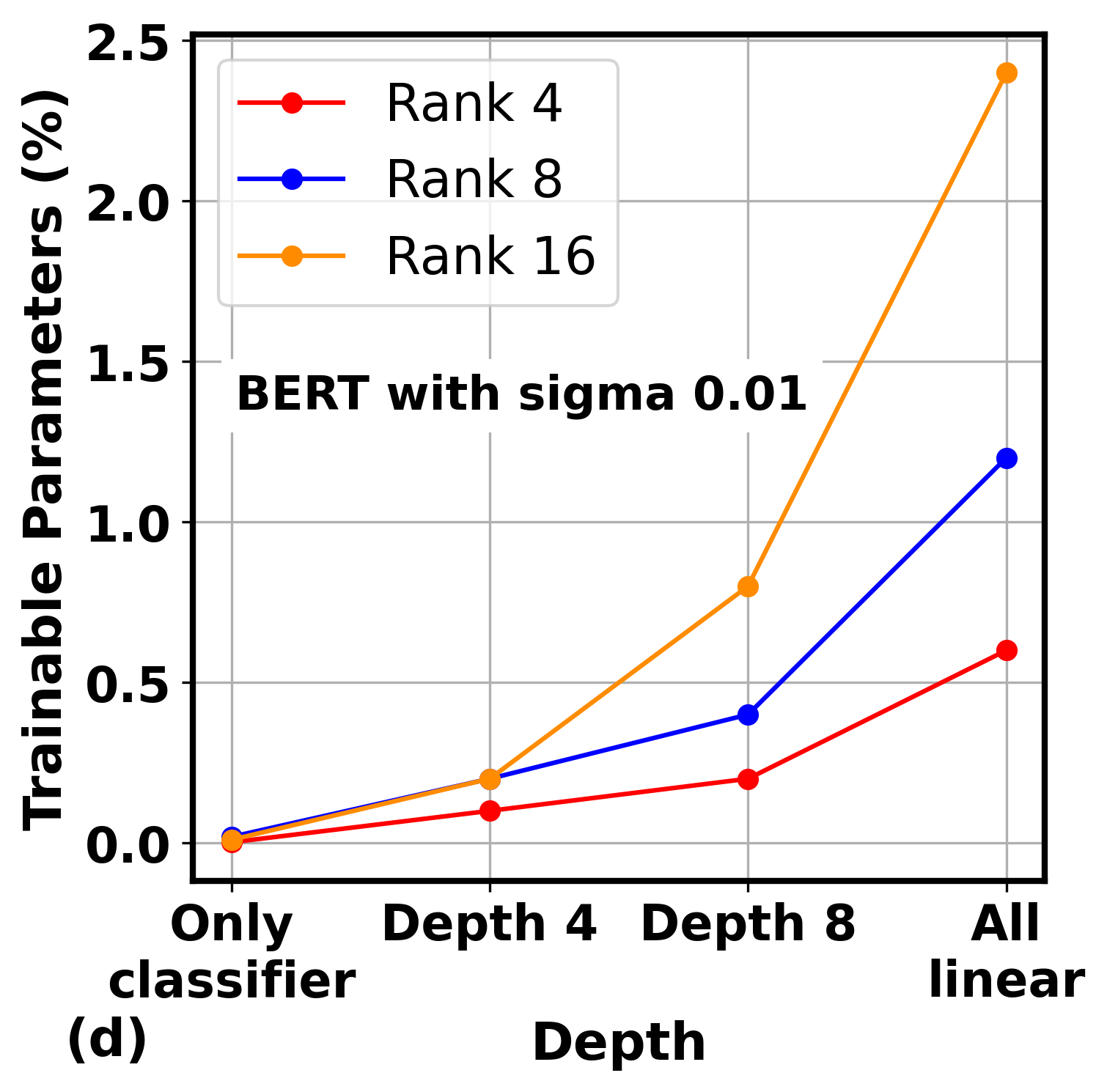}
    \caption{\textbf{(a)} Accuracy degradation with process variability-induced weight perturbations and compensation with low rank tuning. Error compensation at varying \textbf{(b)} rank and \textbf{(c)} depth. \textbf{(d)} Number of surrogate parameters at various rank/depth.\vspace{-15pt}}
    \label{fig:noise_accuracy_drop}
\end{figure}

\subsection{Quantization Error Compensation with Polarized Precision}
Figure 2 shows that accuracy loss from ultra-low-precision processing paths can be effectively mitigated with LoRA-based high-precision surrogate paths. This approach, tested on transformer based models, specifically ViT with CIFAR100 and ImageNet, showed that extreme quantization (e.g., FP4, NF4) caused significant accuracy drops, nearing random guessing. By retraining only low-rank paths layer by layer at varying ranks, we significantly recovered prediction accuracy. Notably, even rank-1 paths effectively compensated for the accuracy loss, with higher ranks (e.g., rank-8) offering further improvements.With accuracy consistently ranging between 74-77\%\ for ImageNet and 73-77\%\ for CIFAR100, even at low ranks and extreme quantization levels (4-bit and 8-bit), LoRA demonstrates its effectiveness in preserving reliable performance.  However, excessively high ranks led to overfitting and diminished accuracy.\vspace{-10pt}

\subsection{Process Variability Resilience with Polarized Precision}
Beyond precision scaling, surrogate paths can also address accuracy loss due to process variability. This is particularly beneficial for analog-domain computing and emerging technologies like memristors~\cite{Xu_Wang_Yan_2021}, PCM~\cite{fantini2020phase}, or MRAM, which integrate processing and storage for enhanced efficiency. However, these technologies are highly sensitive to process variability at advanced nodes, where issues like resistance drift and non-linearity can compromise computational reliability~\cite{velasquez2019error}. Physics-based approaches, such as compute-in-memory, intensify these challenges by closely linking variability to functional accuracy, making systems more vulnerable to hardware imperfections\cite{wan2022accuracy}.

\begin{figure}
    \centering
    \includegraphics[width=0.48\linewidth,height=0.479\linewidth]{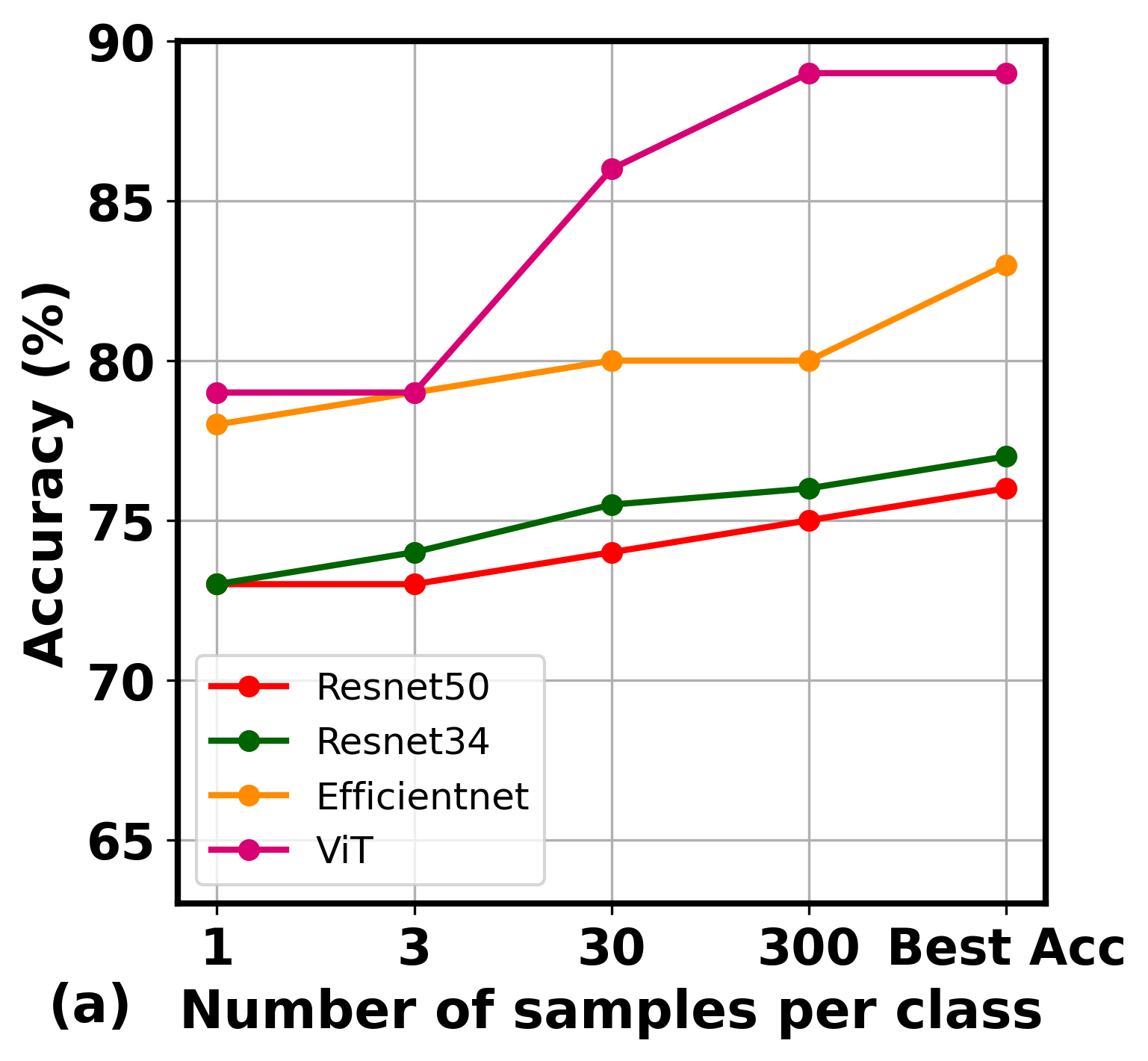}
    \includegraphics[width=0.48\linewidth,height=0.479\linewidth]{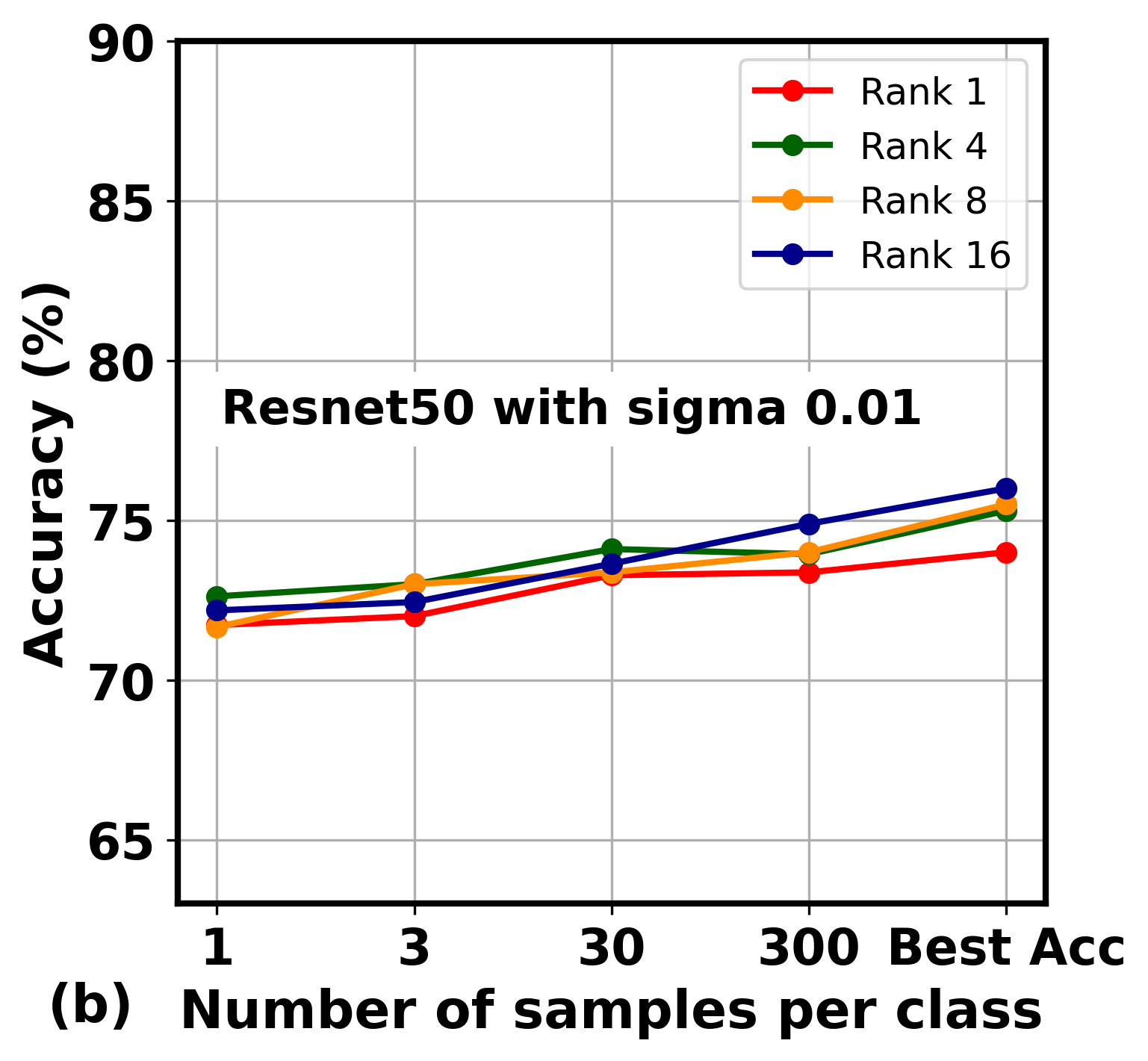} 
    \caption{\textbf{Impact of Sensitivity Ranking and Sample Selection on Accuracy}: {(a)} Sensitivity ranking used to select the most sensitive samples for each class. Accuracy slightly increases with the number of samples. {(b)} Accuracy variation with sample count across different ranks in Resnet50.\vspace{-15pt}}
    \label{fig:accuracy_drop}
\end{figure}

In Fig. 3, we analyze process variability recovery in CNNs and transformer-based models. Fig. 3(a) shows that weight perturbations degrade the accuracy of pretrained models such as ResNet50 and ViT. Weight perturbations were applied to simulate process variability, where weight \( X \) was perturbed as \( X + X \cdot (1 + \mathcal{N}(0, \sigma)) \), with the parameter \( \sigma \) varied. As \( \sigma \) increases, the accuracy of both models decreases. Fig. 3(a) also demonstrates accuracy recovery through low-rank surrogate paths, specifically with rank 4, achieving nearly unperturbed accuracy even when \( \sigma \) is as high as 0.01, i.e. with $~$30\% worst case variations. Fig. 3(b) presents accuracy variation with different ranks under conditions of process variability across several networks while adding the surrogate path only on the linear layers. Notably, for transformer-based models like ViT, accuracy improves significantly with increasing rank, while for BERT, accuracy saturates at rank 8. In contrast, convolutional networks such as EfficientNet show limited improvement with increased rank, indicating less sensitivity to rank tuning.

Fig. 3(c) examines the BERT model's accuracy at various combinations of rank and depth. Accuracy improves considerably when lower-rank adaptations are applied to deeper layers, such as depth 4 or 8, or across all linear layers, yielding substantial improvements. Lastly, Fig. 3(d) illustrates how trainable parameters scale with both depth and rank. The highest accuracy of 82\% was achieved with a BERT model on the IMDB dataset at rank 8, applying LoRA across all linear layers and training only 1.2\% additional parameters.\vspace{-10pt}

\subsection{Minimizing Retraining Data by Sensitivity Ranking}
While low-ranked surrogate paths minimize the parameters required to restore accuracy under ultra-low-precision operations, tuning complexity still depends on the number of samples for retraining. To address this, we developed a sensitivity metric to reduce sample requirements by prioritizing samples with the highest impact on overall prediction loss, minimizing tuning complexity and resource demands. The sensitivity metric computes the loss for each class, back-propagates this loss to assess each parameter's effect on predictions, and calculates a sensitivity score by summing absolute gradient values across all model parameters. This score identifies critical samples that most influence model performance, enabling focused retraining. Specifically, for a model \( f_\theta \) with input samples \( x \) and labels \( y \), the loss \( L_i \) is computed per class, then back-propagated to accumulate gradients for each parameter \( w_j \) as \( sensitivity\_score += \sum |\nabla_\theta w_j| \).

Fig. 4 illustrates model accuracy recovery against process variability (\(\sigma = 0.01\)) when retraining with varying samples per class, showing that models like ResNet50, ResNet34, EfficientNet, and ViT can regain performance even with one sample per class.As sample count increases, accuracy improves significantly, underscoring the effectiveness of retraining with a few carefully selected samples based on the metric. Fig. 4(b) demonstrates the model's resilience under extreme efficiency constraints: rank-1 minimizes training parameters, while a single sample per class maximizes data efficiency. Remarkably, even in this minimal setup, the model sustains strong accuracy, highlighting that sensitivity-based sample selection effectively supports high performance with minimal resources.

\begin{figure*}
    \centering
    \includegraphics[width=0.38\linewidth]{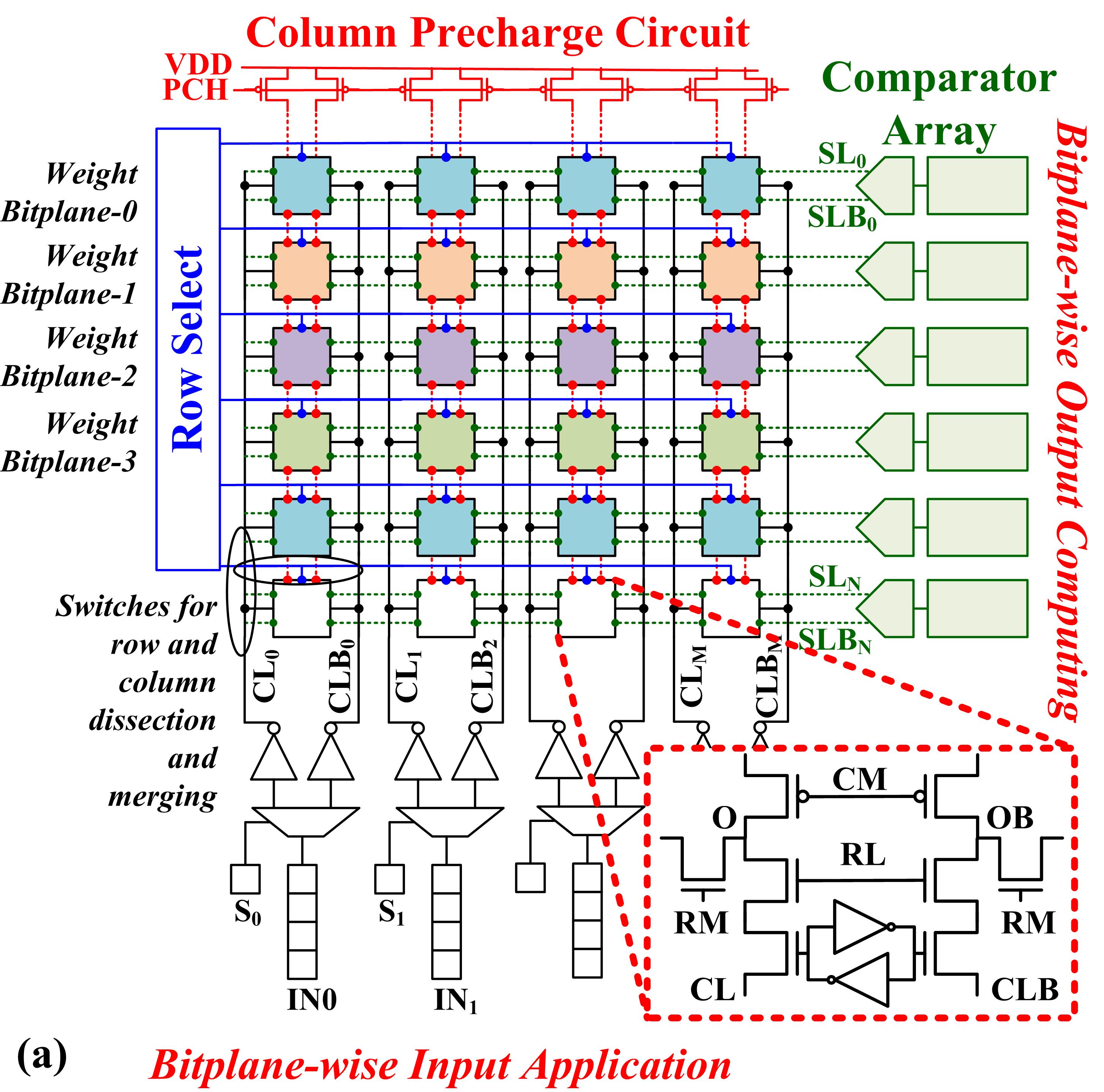}
    \includegraphics[width=0.61\linewidth]{figures/Top_level_arch_cropped.png}
\caption{The NPP approach enhances efficiency by utilizing highly energy-efficient, though potentially unreliable, building blocks for reliable processing. \textbf{(a)} shows an in-memory computing unit with bit-plane-wise separation, removing the need for ADCs/DACs and achieving high energy efficiency through aggressive quantization. To compensate for processing errors, FP32 precision surrogate paths are integrated using digital building blocks, as shown in \textbf{(b)}. \vspace{-15 pt}}

    \label{fig:accuracy_drop}
\end{figure*}

\section{Ultra-Low-Energy Processing with Imprecise Modules with Neural Precision Polarization}
By leveraging extremely low-power, though potentially unreliable, design units, NPP achieves reliable processing through surrogate error compensation paths. Fig. 5 illustrates this potential with a two-component layer implementation, where the first component, an energy-efficient but potentially unreliable processing unit, utilizes bitplane-wise separable in-memory computing scheme from our prior work \cite{darabi2024adc}.
This design achieves full input/output parallelism within the crossbar array, with input vectors applied column-wise and output vectors computed row-wise. The bitplane-wise processing occurs in four steps: (1) precharging bit lines (BL/BLB) and applying the input bitplane, (2) performing parallel local AND operations in SRAM cells for bit-wise products, (3) activating row-merge to sum outputs along sumlines, and (4) comparing sumline values to generate the final output bit. In the inset of Fig. 5(a), the 6T-SRAM-based processing cell is shown for such parallelization with additional ports controlled by the column merge (CM) and row merge (RM) switches to enable full parallel processing. The intermediate nodes O and OB are precharged. When the row line (RL) is activated, depending on input bits at column lines CL/CLB, either O or OB discharges. After this, the RM switches short these lines row-wise, producing the corresponding sum line outputs SL and SLB. The resultant output at the sumlines is compared to generate 1-bit output. For $N$-bit processing, therefore, only $N$ cycles of operations are needed.

Notably, the processing scheme in Fig. 5(a) introduces multiple computational approximations, including staged quantization. Specifically, both input and weight values are quantized to fixed precision, and each bit-plane product-sum is further quantized to a single bit. This approach enhances flexibility by eliminating the need for ADC and DAC operations, significantly improving scalability. However, it is inherently more prone to process variability. Unlike typical in-memory computing units, where operations occur on the more capacitive bit line, computations here are performed on less capacitive nodes due to column separation. Meanwhile, at 8-bit quantization, the memory crossbar requires only $\sim$2.5 fJ energy per MAC operation from our prior work at 16 nm CMOS node \cite{darabi2024adc}. 

To address the reliability limitations of the crossbar, we integrate low-rank floating-point precision paths, which provide compensation for variability and extend precision capabilities. Fig. 5(b) shows the dataflow of surrogate path processing. The scheme processes the floating-point input vector against the stored weight matrix in six steps: \darkgreycircled{1} element-wise exponent addition, \darkgreycircled{2} largest exponent identification ($E_\text{max}$), \darkgreycircled{3} column-wise normalization, \darkgreycircled{4} mantissa scaling by the normalization factor, \darkgreycircled{5} fixed-point scalar products between normalized mantissa and weights, and \darkgreycircled{6} final normalization of the output. However, due to these processing complexities, at FP32, the processing requires $\sim$19 fJ energy per MAC operation at 16 nm CMOS node \cite{xie2015performance}. 

\begin{figure}[t!]
    \centering
    \includegraphics[width=0.46\linewidth,height=0.479\linewidth]{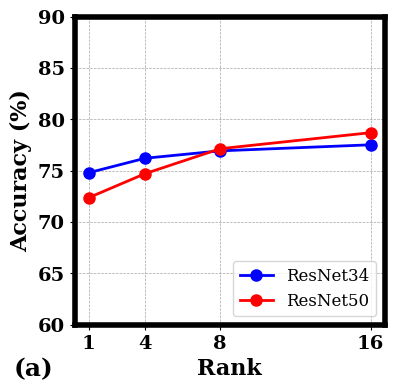}
    \includegraphics[width=0.52\linewidth,height=0.49\linewidth]{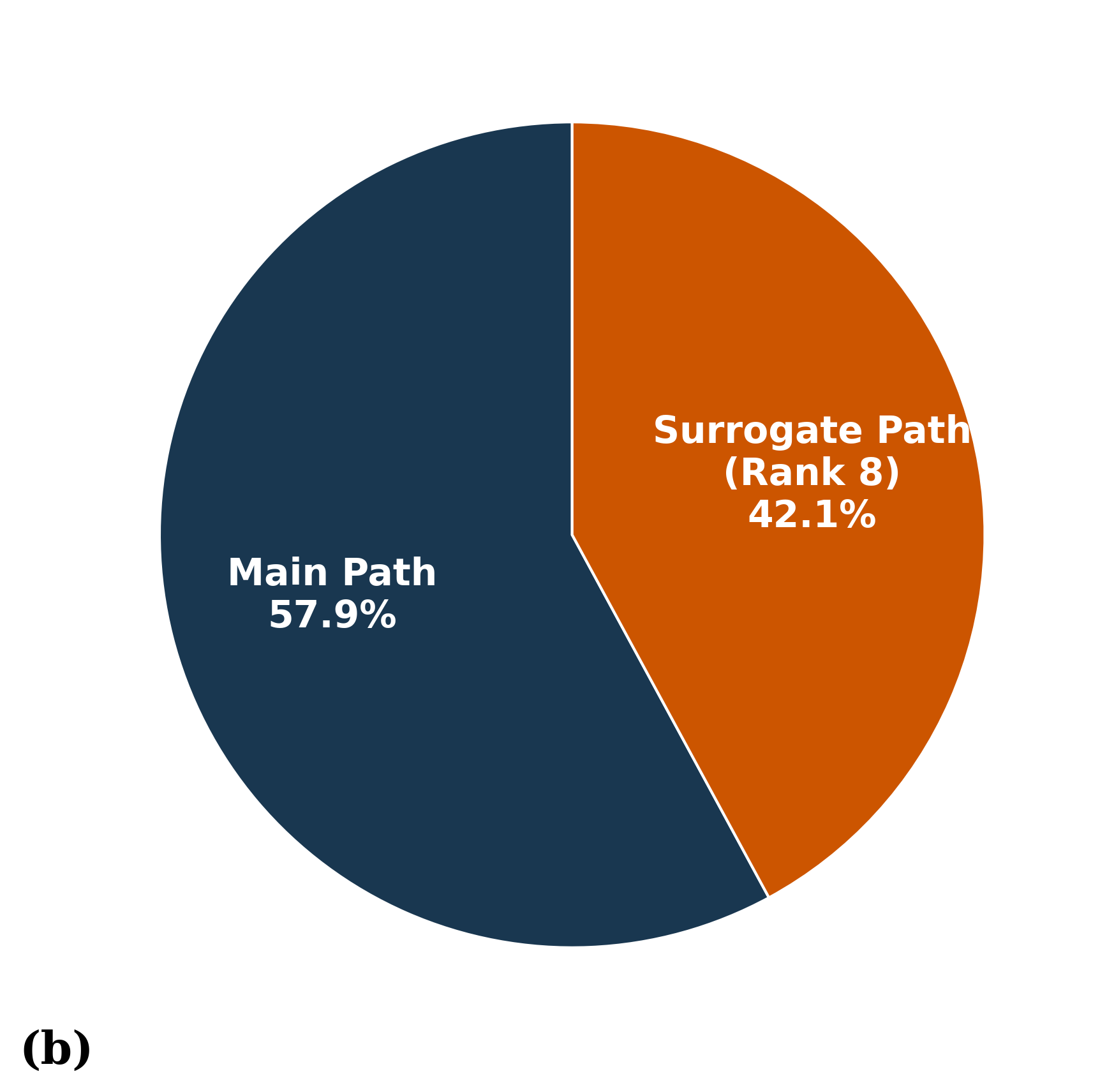}
    \caption{\textbf{Integrated processing of surrogate paths in the last layer:} (a) Results for accuracy recovery with surrogate paths added to the last layer. (b) Energy distribution of the rank-8 surrogate path in the last layer of ResNet-34. \vspace{-15pt}}
    \label{fig:resnet-cim}
\end{figure}

Fig. 6 presents results for integrated processing on ResNet-34 and ResNet-50, where surrogate error compensation paths are added only to the last layer. Notably, rank-8 and rank-16 surrogate paths fully recover accuracy loss despite quantization. Fig. 6(b) shows the energy distribution for the last layer on two processing paths when combining a rank-8 error compensation path with in-memory computing unit, achieving $\sim$464 TOPS/W. Notably, the rank-8 path adds only 0.02\%\ to model parameters while consuming only 42.1\%\ of the energy. The integrated processing approach significantly outperforms prior methods, such as \cite{sinangil20207} which reached 351 TOPS/W with 4-bit precision on 7nm CMOS. Similar benefits can be achieved by integrating the proposed error compensation paths on other compute-in-memory and analog acceleration methods such as \cite{shukla2022mc, nasrin2022enos, nasrin2021mf, nasrin2020supported, shukla2021ultralow, manasi2017skewed, shylendra2020low}. 


\section{Conclusions}
We present a three-pronged approach to drastically reduce resource demands for scalable edge deep learning. First, we lower network precision with low quantization across most layers, compensating accuracy loss via high-precision surrogate paths. Second, to reduce on-device tuning costs, we minimize retraining samples by prioritizing data based on sensitivity. Finally, we demonstrate efficient implementation of low-rank surrogate paths.


\newpage

\bibliographystyle{IEEEtran}
\bibliography{ref.bib}

\end{document}